# EVALUATION OF THE VISUAL ODOMETRY METHODS FOR SEMI-DENSE REAL-TIME

HAIDARA Gaoussou and PENG Dewei

School of Computer Science and Technology, Wuhan University of Technology, Wuhan, China

*ABSTRACT*

*Recent decades have witnessed a significant increase in the use of visual odometry(VO) in the computer vision area. It has also been used in varieties of robotic applications, for example on the Mars Exploration Rovers.*

*This paper, firstly, discusses two popular existing visual odometry approaches, namely LSD-SLAM and ORB-SLAM2 to improve the performance metrics of visual SLAM systems using Umeyama Method. We carefully evaluate the methods referred to above on three different well-known KITTI datasets, EuRoC MAV dataset, and TUM RGB-D dataset to obtain the best results and graphically compare the results to evaluation metrics from different visual odometry approaches.*

*Secondly, we propose an approach running in real-time with a stereo camera, which combines an existing feature-based (indirect) method and an existing feature-less (direct) method matching with accurate semi-dense direct image alignment and reconstructing an accurate 3D environment directly on pixels that have image gradient.*

## 1. INTRODUCTION

The visual odometry occupies an increasingly large and important share of the field of computer vision, navigation, and robotics. Visual odometry is defined as estimation of the trajectory of an object in motion. It was developed in the 1980s to solve the problem of estimating a vehicle's egomotion controlled by a computer and equipped with a TV camera through which the computer can see and run realistic obstacle courses [1]. In addition, it is more useful in situations where other means of localization are not available e.g., GPS, odometry using a motorized wheelchair. The term Visual Odometry was used for the first time in 2004 by Nister et al. as discussed in their paper [2].

The first aim of this paper is to evaluate the performance metrics of two visual odometry methods open source for semi-dense real-time on three public datasets, which are KITTI dataset [3], EuRoC MAV dataset [4], and TUM RGB-D dataset [5], and graphically compare the results. The use of the Umeyama method [6] allowed us to find the best results of performance metrics compared to other results already obtained by certain papers [7, 8, 9, 10， 19].

In order to get best trajectories alignment to the ground truth in closed form, we can implement either the Horn method [11] or the Umeyama method [6]. We used Umeyama method since it has the advantage over the Horn method in obtaining a scale correction that is beneficial for monocular SLAM like LSD-SLAM. It is also a simple method for trajectory alignment to the ground truth in closed form.

We used Umeyama method for estimating minimum value of the mean squared error with two-point patterns $\{x_i\}$ and $\{y_i\}$, $i = 1, 2, 3, …, n$ are specified in $m$-dimensional space to obtain the similarity transformation parameters. They are of the form :

$$e^2(R, t, c) = \frac{1}{n}\sum_{i=1}^{n}\|y_i - (cRx_i + t)\|^2 \qquad (1)$$

where $e^2(R, t, c)$ is the minimum value of the mean squared error of two-point patterns (sets of points), R is the rotation, t is the translation, and c is the scaling.

The Lemma theorem [6] is an improvement of the classical Gauss-Newton method in solving non-linear least squares regression problems. Its solution is applicable to any dimensional problem, though the quaternion method is valid only for point patterns in three-dimensional space. In this paper, secondly, we present a real-time approach with a stereo camera, which is based on the monocular LSD-SLAM [12] which has been combined with ORB-SLAM2 [9] by USB-CAM approach [13] matching with accurate semi-dense direct image alignment and reconstructing an accurate 3D environment directly on pixels that have image gradient (see Figure 1).

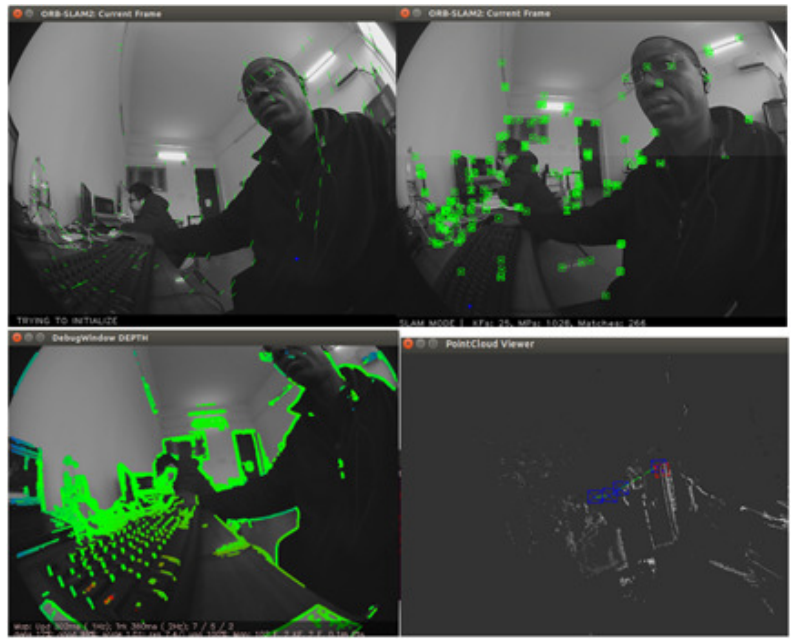

**Figure 1.** Combining Feature-based and Feature-less Method: *Top left* Trying to initialize on ORB-SLAM2. *Top right* Image initialized on ORB-SLAM2. *Bottom left* Keyframes with color-coded semi-dense depth map. *Bottom right* Pose-Graph with semi-dense depth maps.

The rest of the paper is organized as follows: section 2 deals with related work followed by section 3 which is the evaluation of the methods employed, and finally section 4 gives the conclusion.

## 2. RELATED WORK

Visual odometry's visual information is gathered by three methods: Feature Based Methods, Direct Methods, and Hybrid methods. Feature Based Methods process the image to get corners to compare; Direct Methods also compare the entire images to all others to reference them to each other, finding which parts go together; and Hybrid Methods combine these methods. Strengths of feature-based methods are excellent accuracy, robust in dynamic scenes and robust initialization. One weakness of the feature-based methods is low texture areas motion blurs. The strengths of direct methods are robust in low texture areas, under defocus and motion blur, and its dynamic objects strong viewpoint and illumination changes with Rolling-shutter cameras. Weaknesses of the feature-based methods enumerated above are the disadvantages of the direct methods. There are lots of researches done on visual odometry that have been realized using monocular, stereo, and RGD-D sensors [5, 7, 8, 9, 11].

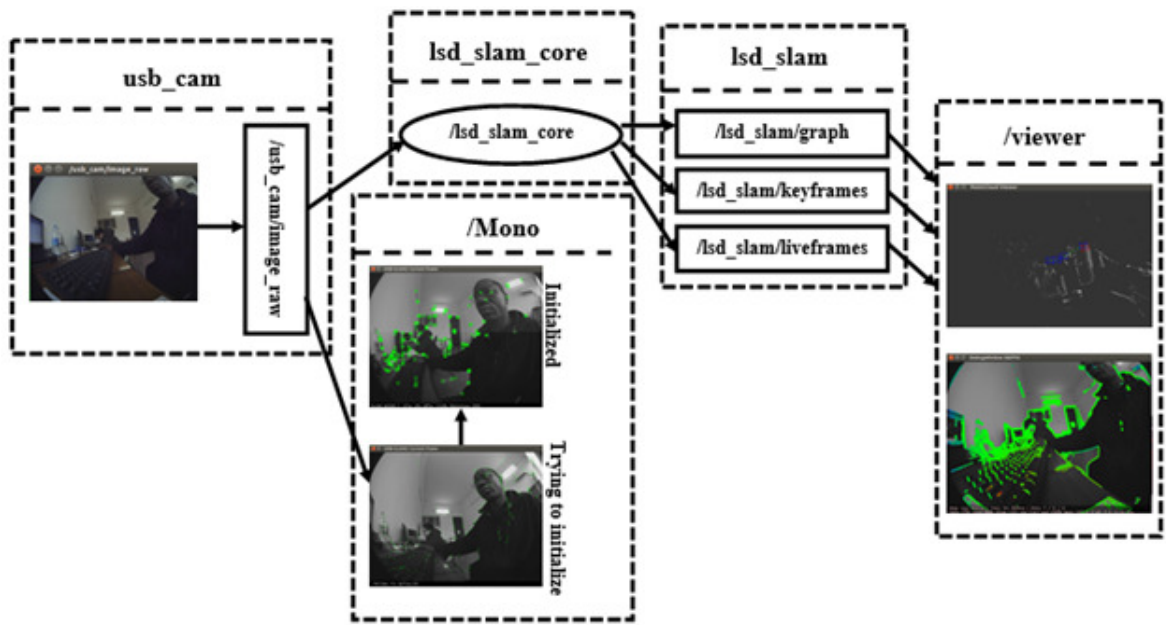

**Figure 2.** Node graph of combining ORB-SLAM2 and LSD-SLAM.

### 2.1. Feature-Based Methods

Feature-based methods are fast to calculate, have good invariance to viewpoint and illumination changes, and can deal well with geometric noise in the system. Raúl et al. [8] proposed a SLAM family with all the properties in the PTAM known as ORB-SLAM. In addition, new features are also added such as the computing of the camera trajectory in real-time and a variety of environments with sparse 3D reconstruction (e.g. from car, outdoor and indoor scenes), allowing wide baseline loop closing and global relocalisation in real-time, and including full automatic initialization. This year, they have recently extended ORB-SLAM to a complete SLAM system that can perform reuse in real time map it on CPUs [9]. [8, 9, 14, 15], used ORB features, a visual detector which is able to detect and report the basic features of an image (for example shape, color, and so on). In the context of camera pose estimation in an unknown scene, Davison et al. [16] have proposed an approach that use a sparse map of high quality features, while Klein et al. [17], proposed an approach that uses a much denser map of lower-quality features.

S-PTAM [18] is one of the best point-based sparse methods to calculate the real-time camera trajectory, has been heavily used to solve the parallel nature of the SLAM problem in real scale of the environment in the similar scenario as [24]. While S-PTAM was originally introduced for monocular systems and Standalone variants have meanwhile been developed. Therefore, the stereo setting provides a metric 3D map reconstruction for every frame of the stereo images, enhancing the accuracy of the mapping process with regard to monocular SLAM and preventing the bootstrapping problem.

## 2.2. Direct Methods

In contrast, Direct methods register images to each other using correlation, which does not depend on finding specific repeatable features and instead matches the image as a whole. Engel et al. [12] implemented a method to SLAM by proposing a direct monocular camera algorithm without extraction of feature. The inverted and normalized depth map does not detect scale shift from one frame to another because, this requires sufficient camera translation. LSD-SLAM [12] only uses TUM RGB-D dataset. [12] later extended that work by introducing stereo cameras, which directly succeeded in reconstructing a map of the environment at the same time as the trajectory using photometric and geometric residuals, and eliminates scale uncertainties and complexities with degenerate motion alongside the sight line [19]. Besides the TUM RGB-D dataset use, [19] also run on KITTI dataset and EuRoC MAV dataset. Stumberg et al. [20] recently presented a monocular method for tracking the camera motion and getting an environment with semi-dense reconstruction running in real-time using LSDSLAM. [21] used a stereo camera pair and the quadrifocal tracking methodology's procedure for detecting and tracking the 6DOF trajectory with an accurate, robust and efficient 3D scene reconstruction. Pizzoli et al. [22] solved the estimation problem to calculate a probabilistic, monocular and accurate dense 3D scene reconstruction observed by a single moving camera in real time. In their work known as Stereo Direct Sparse Odometry (Stereo DSO), Wang et al. [23] used quantitative and qualitative evaluations on the KITTI and Cityscapes datasets to achieve good real-time performance for tracking a stereo camera than Stereo ORB-SLAM2.

[24] proposed a Direct point-based rendering method for visualization of point-based data by combining volume and point-based renderings in GPU hardware developed under the name of General Purpose Computing on GPU (GPGPU) where several algorithms with higher computational difficulty have already been competently parallelized.

Comparable all sensors, cameras also require calibration for rectifying the distortions in the camera images because of the internal parameters of the camera and for getting the world coordinates from the camera coordinates. Recently an online calibration of the auto exposure video has been proposed by Bergmann et al [25] in order to estimate for the vignetting, photometric response function, and exposure times. This provides a robust and accurate photometric calibration for arbitrary video sequences and meaningfully improves the direct VO methods quality.

## 2.3. Hybrid Methods

The idea of the hybrid method is to benefit from the different strengths of Feature-based and Direct approaches. [26] demonstrated a direct and sparse formulation for SfM (Structure from Motion), which combines the advantage of the different strengths of direct methods with the flexibility of sparse approaches. Recently, Krombach et al. [7] have presented a hybrid method for combining feature-based tracking which directly aligns images with semi dense map. [27] introduced a semi-direct monocular approach that eliminates the feature extraction use and matching methods for motion estimation by combining high framerate motion estimation with probabilistic mapping method.

In contrast to [7], our approach is much closer to the ground truth provided by KITTI dataset and accumulates very small drift (see Figure 3). Looking at the two trajectories obtained from our method (estimated trajectory in colored solid line and ground in gray dash-dotted), you would think that it is only a trajectory instead of two. That clearly demonstrate the robust and efficient 3D scene reconstruction of the Umeyama method.

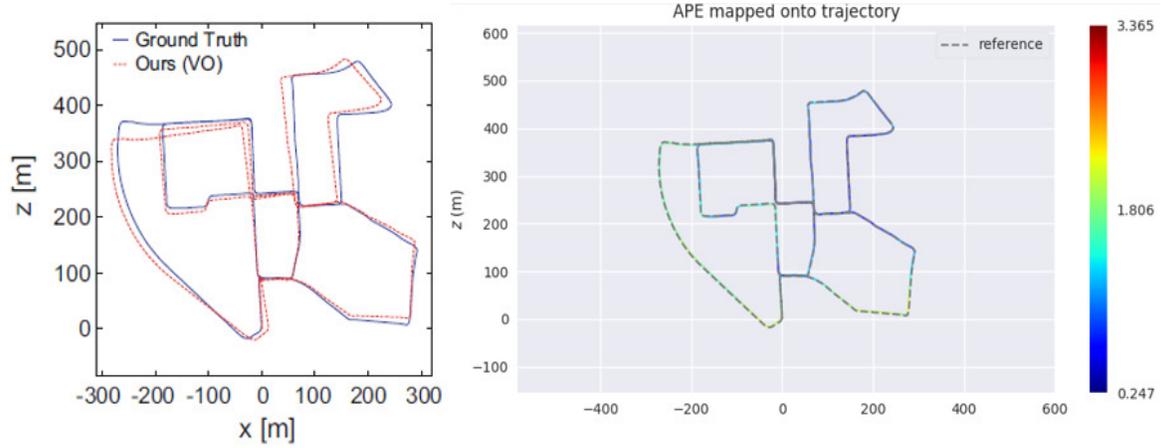

**Figure 3.** Comparison between the hybrid odometry proposed by [7] (left) and our method (right) on Sequence 00 with ground truth provided by KITTI dataset.

## 2.4. Combining ORB-SLAM2 and LSD-SLAM

ORB-SLAM2 is a complete SLAM system in real-time that runs the approach in the TUM dataset as monocular or RGB-D, in the EuRoC MAV dataset as monocular or stereo, in the KITTI dataset as monocular or stereo, and ROS node to treat RGB-D streams, stereo or live monocular [9]. It consists of receiving the images as input and executing 3 parallel threads: Tracking which allows to track the camera from the images, Local Mapping that aims to process new reference images, and Loop Closing which searches all the new images added if there is a loop-closure. Pose Optimization of ORB-SLAM2 uses Motion-only bundle adjustment by optimizing camera orientation $R$ and position $t$, and minimalizing error between similar 3D points in world coordinates and key points. In order to make that, it uses the Levenberg–Marquardt method applied in g2o [28].

$$\{R, t\} = arg\min_{R,t} \sum_{i \in \chi} \rho \left( \left\| X^i_{(\cdot)} - \pi_{(\cdot)}(RX^i + t) \right\|^2_\Sigma \right) \quad (2)$$

where the covariance matrix associated with the key-point scale is $\Sigma$, the robust Huber cost function is $\rho$. The monocular function $\pi_m$ and rectified stereo function $\pi_s$ are the projection functions $\pi_{(\cdot)}$, and they are given by:

$$\pi_m\left(\begin{bmatrix} X \\ Y \\ Z \end{bmatrix}\right) = \begin{bmatrix} f_x \frac{X}{Z} + c_x \\ f_y \frac{Y}{Z} + c_y \end{bmatrix} \quad (3) \qquad \pi_s\left(\begin{bmatrix} X \\ Y \\ Z \end{bmatrix}\right) = \begin{bmatrix} f_x \frac{X}{Z} + c_x \\ f_y \frac{Y}{Z} + c_y \\ f_x \frac{X-b}{Z} + c_x \end{bmatrix} \quad (4)$$

where the focal lengths are $f_x$ and $f_y$, and the principal points are $c_x$ and $c_y$.

LSD-SLAM [12] is a direct approach to SLAM with a monocular camera running in real-time on CPU and even on a smartphone if it is used only in odometry task. It only uses TUM dataset, and therefore it can't work with KITTI dataset and EuRoC MAV dataset for three reasons: first, it is low frame-rate, hence very fast inter-frame motion, secondly, there are strong rotations without much translation during turns and thirdly, it has a relatively small field of view, and forward-motion. Finally, to solve these problems, a novel method Stereo-LSD-SLAM is proposed by Engel et al. [19]. The second and third problems can be solved by using stereo-SLAM and the first can be offset by using a somewhat meaningful motion model (e.g. assuming constant motion) to initialize tracking. In this paper, we utilize the LSD-SLAM open source release and Stereo LSD-SLAM [19]. There are three principal components that make LSD-SLAM: tracking, depth map estimation, and map optimization. The tracking consists in continuously tracking new images from the camera, and estimates the rigid position using the current default keyframe position. The depth map estimation consists in using the frames retrieved by the tracking phase to refine or replace the recent keyframe. The map optimization is defined and once a key-frame has been replaced by a new one, this one will not be refined anymore and can thus be incorporated in the global map, and the optimization will seek to detect loops of loops with other frames. Mathematical optimization appears in many forms in the algorithms including Golden Section Method that we used for calculating of absolute trajectory error (ATE) performed by LSD-SLAM on TUM RGB-D dataset.

$$f: \mathbb{R} \rightarrow \mathbb{R} \quad (5)$$

The Golden Section Method is a theorem to solve one dimensional optimizations problems of form $\mathbb{R} \rightarrow \mathbb{R}$ over the interval [a, b]. The function $f$ has to be convex in the range. In contrast to many other optimization techniques, the function is solely evaluated and the gradient or Hessian is not required.

The experiment was carried out in the Laboratory. The camera was calibrated in order to offer a good accuracy before it was used. USB_CAM [13] allowed LSD-SLAM [12] and ORB-SLAM2 [9] to display the raw image at a very high frame rate from its node (usb_cam_node). LSD-SLAM initialized each of point in a plane, while ORB-SLAM2 waited up till there was sufficient parallax, initializing correctly from the fundamental matrix. When ORB-SLAM2 tried to initialize, the camera was moved a bit in the image plane to initialize SLAM (Figure 3.). Therefore, LDS-SLAM uses everything that's colored in the room and everything that has a gradient which gives a much denser and nicer map.

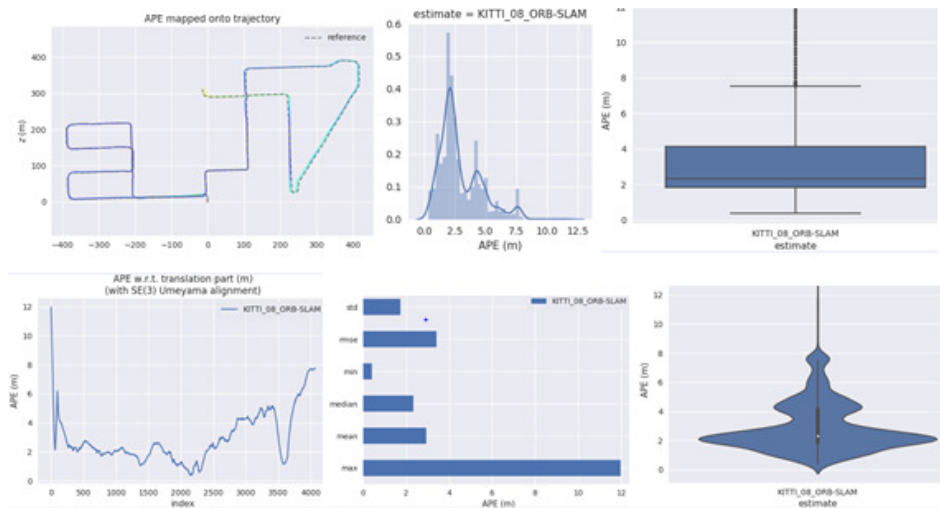

**Figure 4.** Comparing multiple result files from the metrics on sequence 08 of KITTI dataset. *Top left, top right* and *bottom left* plot are the results of map, box plot and raw of poses from the ORB-SLAM2 on the sequence 08 of KITTI dataset and the ground truth (reference). *Top middle* and *bottom right* print information and statistics of histogram and violin histogram poses from the ORB-SLAM2 on the sequence 08 of KITTI dataset and the ground truth. *Bottom middle* saves the statistic poses in a table from the ORB-SLAM2 on the sequence 08 of KITTI dataset and the ground truth.

## 3. EVALUATION

For performance evaluation metrics, we evaluated and tested different methods referred to above on the same datasets. We run ORB-SLAM2 [9] and LSD-SLAM [12] in an Intel Core™ i5-3317U laptop computer with 12GB RAM using Ubuntu 14.04 (trusty) and ROS indigo. We evaluated the Absolute Trajectory Error (ATE) by Root Mean Squared Error (RMSE) by using the estimated trajectory to compare with the ground-truth data provided by dataset. Calculating the error of the absolute trajectory performed by LSD-SLAM on TUM RGB-D dataset is difficult due to two factors: scale ambivalence, LSD-SLAM uses only one camera for input, and also has no depth input; and alignment, LSD-SLAM does not get a fixed initial pose along with the scale ambivalence [ 29].  In order to resolve the problem of that two factors, before determining the ATE, we made an alignment algorithm performed by Umeyama method [6] and extended the evo_ape [30] to obtain the optimum scale, the Golden Section Method is referenced in section 2.4. While calculating the ATE performed by ORB-SLAM2 on different datasets, we used the corresponding poses directly and compared between estimate and ground truth given a pose relation by using evo_ape.

**Table 1.** Absolute Trajectory Error (RMSE) results performed by LSD-SLAM and ORB-SLAM2 on KITTI Dataset. The symbol '- ' denotes situations where LSD-SLAM could not run on KITTI dataset and the symbol 'x' indicates that this paper doesn't use the dataset. RMSE results LSD-SLAM of [9, 19] on the sequences of KITTI dataset are the same.

| KITTI Dataset | ORB-SLAM2 in meter | | | | LSD-SLAM in meter | | | |
|---|---|---|---|---|---|---|---|---|
| | Our | [7] | [9] | [19] | Our | [7] | [9] | [19] |
| 00 | 1.25 | 8.30 | 1.3 | x | - | - | **1.0** | **1.0** |
| 01 | **10.09** | 335.52 | **10.4** | **x** | - | - | **9.0** | **9.0** |
| 02 | 6.54 | 18.66 | 5.7 | x | - | - | **2.6** | **2.6** |

| 03 | 0.78 | 11.91 | 0.6 | x | - | - | **1.2** | **1.2** |
|---|---|---|---|---|---|---|---|---|
| 04 | **0.26** | **2.15** | **0.2** | x | - | - | **0.2** | **0.2** |
| 05 | 0.75 | 4.93 | 0.8 | x | - | - | **1.5** | **1.5** |
| 06 | 0.77 | 16.01 | 0.8 | x | - | - | **1.3** | **1.3** |
| 07 | 0.55 | 4.30 | 0.5 | x | - | - | **0.5** | **0.5** |
| 08 | 3.40 | 38.80 | 3.6 | x | - | - | **3.9** | **3.9** |
| 09 | 2.91 | 7.46 | 3.2 | x | - | - | **5.6** | **5.6** |
| 10 | 1.04 | 8.35 | 1.0 | x | - | - | **1.5** | **1.5** |
| Mean | **2.58** | **41.49** | **2.55** | **x** | **-** | **-** | **2.57** | **2.57** |

## 3.1. Accuracy

The KITTI benchmark [3] consists of 22 stereo sequences with maximum speed of 80 km/h, and resolution of 1241×376 pixels, whose 11 sequences (00-10) with ground truth trajectories for exercise was used in our experiments. The testing results for ATE(RMSE) are shown in the Table 1 and the visualization comparing multiple result files from the metrics on sequence 08 in shown in Figure 4. None of LSD-SLAM trials on all sequences (00 to 10) of the KITTI dataset was executed successfully, because LSD-SLAM can only run on the TUM RGB-D dataset and the reasons mentioned in the section 2.4. By observing the results of the ATE from sequences of KITTI dataset in Table1, we note that the highest value ATE is the sequence1 as also stated by [7]. To calculate ATE, the estimated trajectory was first aligned onto the ground truth trajectory and this can be done by using the Umeyama method [6] (formula (1)). Based on the formula (1), when the rotation and the translation increase, the result in formula (1) also increases. In addition to the argument of [7], the values of the rotation and translation matrices of the sequence 01 are highest on all sequences (00 to 10) from KITTI dataset, and the ATE value of the sequence 01 is therefore highest. The values of the rotation and translation matrices of the sequence 04 are the lower values, with the same reasoning, the ATE value of the sequence 04 has to get the lower value. Figure 4 summarizes the comparing multiple result files from the metrics on sequence 08 of KITTI dataset. The top left, top right and bottom left plots are the results of map, box plot and raw of poses from the ORB-SLAM2 on the sequence 08 of KITTI dataset and the ground truth (reference). The top middle and bottom right print information and statistics of histogram and violin histogram are from the ORB-SLAM2 on the sequence 08 of KITTI dataset and the ground truth. The bottom middle saves the statistic poses in a table from the ORB-SLAM2 on the sequence 08 of KITTI dataset and the ground truth.

**Table2:** ATE Results performed by LSD-SLAM and ORB-SLAM2 on EuRoC MAV Dataset. The symbol '- ' denotes situations where LSD-SLAM could not run on KITTI dataset and the symbol 'x' indicates that this paper doesn't use the dataset.

| EuRoC | LSD-SLAM in meter | | | | ORB-SLAM2 in meter | | | |
|---|---|---|---|---|---|---|---|---|
| Dataset | Our | [7] | [9] | [10] | our | [7] | [9] | [10] |
| MH_01 | - | x | - | x | **0.038** | x | 0.035 | 0.075 |
| MH_02 | - | x | - | x | 0.044 | x | **0.018** | 0.084 |
| MH_03 | - | x | - | x | 0.043 | x | **0.028** | 0.087 |
| MH_04 | - | x | - | x | 0.144 | x | **0.119** | **0.217** |
| MH_05 | - | x | - | x | 0.088 | x | **0.060** | 0.082 |
| **Mean** | **-** | **x** | **-** | **x** | **0.07** | **x** | **0.052** | **0.109** |
| | | | | | | | | |
| V1_01 | - | **0.19** | 0.066 | x | 0.088 | 0.79 | 0.035 | **0.027** |
| V1_02 | - | **0.98** | 0.074 | x | 0.064 | 0.98 | 0.020 | 0.028 |

| V1_03 | - | - | **0.089** | x | 0.077 | **2.12** | 0.048 | - |
| V2_01 | - | 0.45 | - | x | 0.063 | **0.50** | 0.037 | 0.032 |
| V2_02 | - | 0.51 | - | x | 0.061 | 1.76 | 0.035 | 0.041 |
| V2_03 | - | - | - | x | **0.161** | - | - | 0.074 |
| **Mean** | - | 0.53 | 0.08 | x | 0.09 | 1.23 | 0.035 | 0.04 |

The EuRoC MAV Dataset [4] includes 11 stereo sequences of the visual-inertial datasets whose 5 sequences are recorded in a large machine hall provided by WVGA images at 20 Hz and 752x480 pixels, and 6 sequences with two easy, two medium and two difficult, recorded in a Vicon room equipped with a motion capture system. Tables 2 reports the ATE Results performed by LSD-SLAM and ORB-SLAM2 on EuRoC MAV Dataset. Unlike ORB-SLAM2 which loses track for the difficult trajectories V1_03 and V2_03 [7, 9, 10], we could run on all the sequences of the EuRoC MAV dataset to successfully find the tracking and calculate the absolute trajectory error (RMSE) to get the best results in most cases. The Figure 5 shows the accuracy and closer form of the Umeyama Method alignment on the different sequences of the EuRoC MAV dataset.

Table 3. ATE Results performed by LSD-SLAM and ORB-SLAM2 on TUM RGB-D Dataset. The symbol '- ' denotes situations where LSD-SLAM could not run on KITTI dataset and the symbol 'x' indicates that this paper doesn't use the dataset.

| TUM RBG-D Dataset | LSD-SLAM | | | ORB-SLAM | | |
|---|---|---|---|---|---|---|
| | Our | [8] | [9] | our | [8] | [9] |
| FR1_360 | 0.07 | x | x | **0.24** | x | x |
| FR1_xyz | **0.01** | 9.00 | x | **0.01** | 0.90 | x |
| FR1_rpy | 0.03 | x | x | 0.02 | x | x |
| FR1_desk2 | 0.04 | x | x | 0.02 | x | 0.022 |
| FR1_desk | 0.03 | 10.65 | x | **0.01** | 1.69 | 0.016 |
| FR1_plant | 0.06 | x | x | **0.01** | x | x |
| FR1_floor | 0.04 | **38.07** | x | 0.02 | 2.99 | x |
| FR1_room | 0.10 | x | x | **0.05** | x | 0.047 |
| FR1_teddy | **0.11** | x | x | 0.04 | x | x |
| **Mean** | 0.05 | 19.24 | x | 0.05 | 1.86 | 0.17 |

The TUM RGB-D Dataset [5] is composed of great data sequences set within an office and an industrial room containing both RGB-D data from the Kinect and ground truth from the capture system of motion. Table 3 summarizes the ATE results over the 9 training sequences of TUM RGB-D dataset, and it was found out that all values obtained from ATE on ORB-SLAM2 are better than those obtained from ATE on LSD-SLAM, except from FR1_360. Every row of TUM RGB-D dataset trajectory format has 8 entries comprising timestamp (in seconds), orientation and position (as quaternion) with very value separated by a space. Based on the association of the estimated poses with ground truth poses using the timestamps, we aligned the estimated trajectory to the ground truth using singular value decomposition and calculated the difference between all pairs of poses. The results showed 745 absolute pose pairs calculating APE for translation over LSD-SLAM and 744 on ORB-SLAM, and the values of the translation matrices of the sequence Fr1_360 on LSD-SLAM are lower than those of ORB-SLAM. By applying the formula (1), the values of the matrix of the translation are higher and the result of the formula (1) is the highest, and this demonstrate the ATE value of FR1_360 from ORB-SLAM is higher than that of LSD-SLAM. Figure 6 shows the visualizations of the absolute ATE on the FR1_floor sequence of LSD-SLAM and on the FR1_floor sequence of ORB-SLAM, and the comparison of the ATE on the FR1_floor sequence of LSD-SLAM, the ATE on the FR1_floor sequence of ORB-SLAM with the ground truth. That shows the ATE on the FR1_floor sequence of LSD-SLAM is much closer and less drift with the ground truth. In Table 3, we compare our results to [8, 9] and our results on all sequences are better than [8, 9] except FR1_rom.

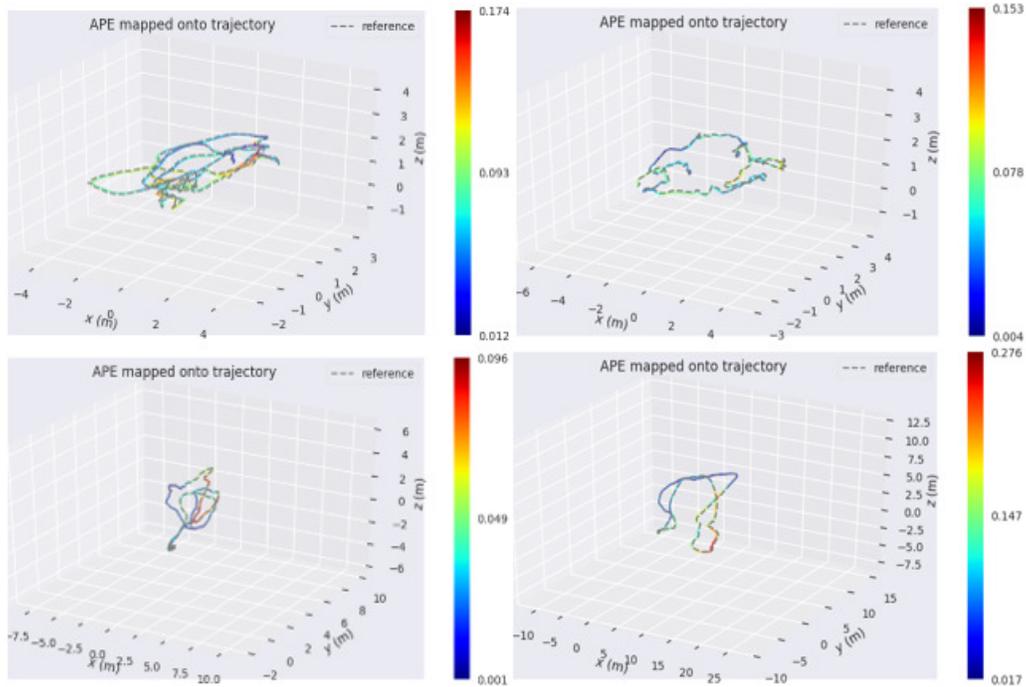

**Figure 5.** Estimated trajectory (colored) and ground truth (gray) on V1_01_easy (top left), V2_01_easy (top right) MH_02_medium (bottom left) and MH_04_difficult (bottom right) from the EuRoC MAV dataset.

### 3.2. Runtime

As [7], we also evaluated the average runtime of two popular methods onto three public datasets that we used in this paper. Table 4 reports the timing results on three public datasets with image resolutions by default. The average tracking time per frame is below the inverse of the camera framerate for all sequences as proposed by [9]. We noted the timing results of the tests of the same sequence from ORB-SLAM2 are invariable while those of LSD-SLAM are variable, and we took the results of the third test of each sequence. Therefore, our timing results on the three public datasets are better than the results already obtained by [7].

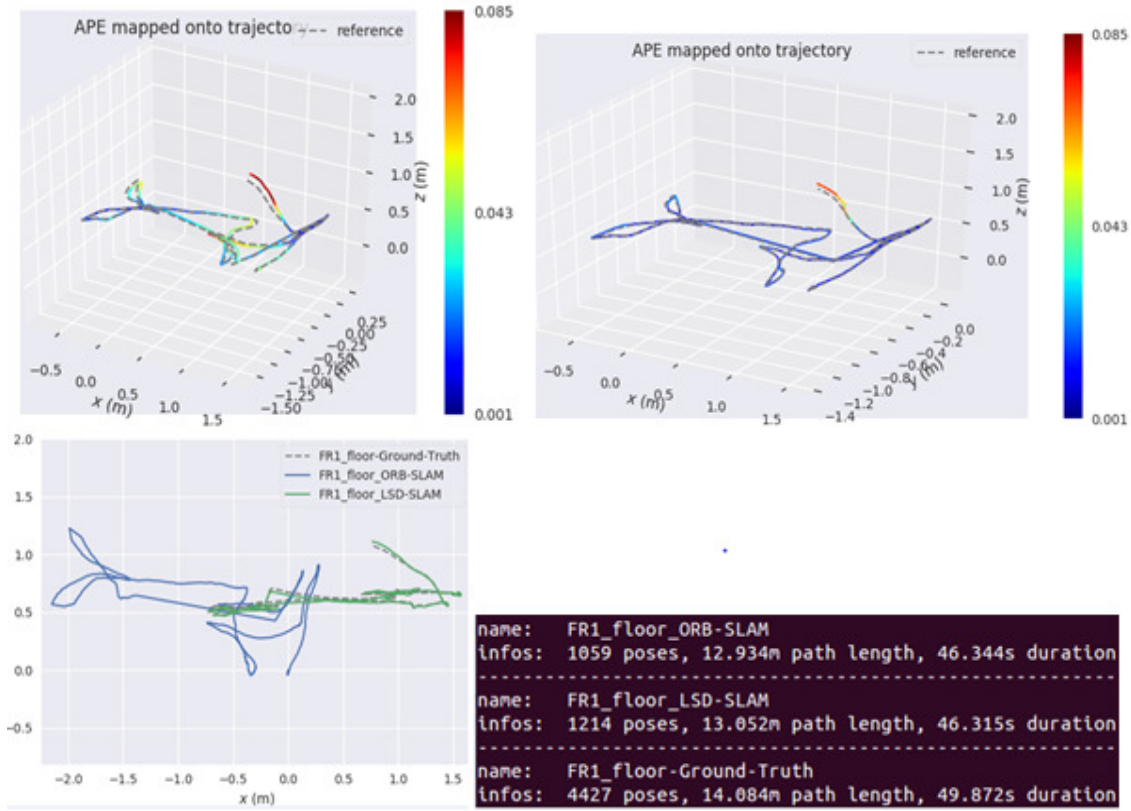

**Figure 6.** *Top left* Plot of TUM pose file on LSD-SLAM and the ground truth(reference). *Top right* Plot of TUM pose file on ORB-SLAM and the ground truth. *Bottom left* Plots of TUM pose files on ORB-SLAM and LSD-SLAM, and the ground truth. *Bottom right* Plots information of TUM pose files on ORB-SLAM and LSD-SLAM, and the ground truth.

## 4. CONCLUSIONS

VO has been used in a lot of research over the past two decades. There has been a lot of progress which is due in no small part to new technologies such as sophisticated sensors, powerful CPU, large amounts of RAM and many more. With the use of the Umeyama method, a closed-form solution of the least-squares problem of the similarity transformation parameter estimation for point patterns in 3D space allowed to find the best alignment of the estimated trajectory with the ground truth trajectory. In most cases, our evaluation metrics on three popular public datasets shows that our results achieved the state-of-the-art accuracy and runtime compared to other results already obtained by certain papers [7, 8, 9, 10]. Unlike [7, 9, 10], we could run on all the sequences of the EuRoC MAV dataset to find the tracking and calculate the absolute trajectory error (RMSE) with success.

The study was based on the monocular LSD-SLAM [12] which was combined with ORB-SLAM2 [9] by USB-CAM approach [13] using a stereo camera, corresponding to the direct images alignment by using geometric and photometric residuals on a set of semi-dense pixels, and a very precise 3D reconstruction of the environment. The goal of our approach is to run in parallel a large number of executables that can exchange information synchronously to further specify the results of LSD-SLAM by a stereo camera.

**Table 4.** Average runtimes on two methods used. The symbol '- ' denotes situations where LSD-SLAM could not run on KITTI dataset, and EuRoC MAV dataset.

| Dataset | Method | Tracking in Milliseconds |
|---|---|---|
| KITTI dataset | LSD-SLAM | - |
| | ORB-SLAM | 7.447804 |
| EuRoC MAV dataset Vicon Room | LSD-SLAM | - |
| | ORB-SLAM | 1.637305 |
| EuRoC MAV dataset Machine Hall | LSD-SLAM | - |
| | ORB-SLAM | 2.035506 |
| TUM RGB-D dataset | LSD-SLAM | 28.2222222 |
| | ORB-SLAM | 1.9250758 |

## ACKNOWLEDGEMENTS

This work was supported by the China Scholarship Council (CSC). We would like to thank Arouna KONATE and Tina WEMEGAH for their supports and helpful discussions.

**AUTHORS**

HAIDARA Gaoussou is Master Student Candidate in Computer Science and Technology, School of Computer Science and Technology from Wuhan University of Technology, Wuhan, China. I did primary and high school, and part of the university in Mali. His research concentrates on Computer Vision and Augmented Reality.

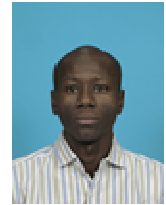

PENG Dewei obtained a PhD in Computer Science and Technology from the Wuhan University in 2004, China. His research concentrates on Mobile Internet, and Intelligent methods and techniques. Until 2009 he was associate professor at the Department of computer science, Wuhan University of Technology, China. He has published more than 8 papers on these topics.

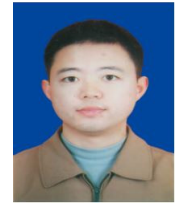